\documentclass[11pt]{article}

\usepackage{emnlp2021}

\usepackage{times}
\usepackage{latexsym}

\usepackage[T1]{fontenc}

\usepackage[utf8]{inputenc}

\usepackage{microtype}
\usepackage{amsmath, amssymb}
\usepackage{booktabs}
\usepackage{graphicx}
\usepackage{multirow}
\usepackage{subfigure}
\usepackage{algorithm}
\usepackage{algorithmic}
\usepackage{bbm}

\newcommand\blfootnote[1]{%
\begingroup
\renewcommand\thefootnote{}\footnote{#1}%
\addtocounter{footnote}{-1}%
\endgroup
}

\title{Exploiting Global and Local Hierarchies for Hierarchical Text Classification}

\author{
    Ting Jiang\textsuperscript{\rm 1}, Deqing Wang\textsuperscript{\rm 1,4,$*$}, Leilei Sun\textsuperscript{\rm 1}, \\{\bf Zhongzhi Chen\textsuperscript{\rm 2},  Fuzhen Zhuang\textsuperscript{\rm 1,3,4}, Qinghong Yang\textsuperscript{\rm 2}} \\
    \textsuperscript{\rm 1}SKLSDE, School of Computer, Beihang University, Beijing, China \\
    \textsuperscript{\rm 2}School of Software, Beihang University, Beijing, China\\
    \textsuperscript{\rm 3}Institute of Artificial Intelligence, Beihang University, Beijing, China\\
    \textsuperscript{\rm 4}Zhongguancun Laboratory, Beijing, China\\
    \{royokong, dqwang, leileisun, jongjyh, zhuangfuzhen, yangqh\}@buaa.edu.cn\\
}

\begin{document}
\maketitle

\blfootnote{ $*$ Corresponding Author.}
\begin{abstract}
  Hierarchical text classification aims to leverage label hierarchy in multi-label text classification.
  Existing methods encode label hierarchy in a global view, where label hierarchy is treated as the static hierarchical structure containing all labels.
 Since global hierarchy is static and irrelevant to text samples,  it makes these methods hard to exploit hierarchical information.
 Contrary to global hierarchy,
 local hierarchy as a structured labels hierarchy corresponding to each text sample. It is dynamic and relevant to text samples, which is ignored in previous methods.
To exploit global and local hierarchies,
we propose Hierarchy-guided BERT with Global and Local hierarchies (HBGL), which utilizes the large-scale parameters and prior language knowledge of BERT to model both global and local hierarchies.
Moreover,
HBGL avoids the intentional fusion of semantic and hierarchical modules by directly modeling semantic and hierarchical information with BERT.
Compared with the state-of-the-art method HGCLR,
our method achieves significant improvement on three benchmark datasets.
Our code is available at \url{http://github.com/kongds/HBGL}.

\end{abstract}

\section{Introduction}

Hierarchical text classification (HTC) focuses on assigning one or more labels from the label hierarchy to a text sample~\cite{sun2001hierarchical}.
As a special case of multi-label text classification, HTC has various applications such as news categorization~\cite{kowsari2017hdltex} and scientific paper classification~\cite{lewis2004rcv1}.
The methods in HTC aim to improve prediction accuracy by modeling the large-scale, imbalanced, and structured label hierarchy~\cite{mao2019hierarchical}.



To model the label hierarchy, recent methods~\cite{zhou2020hierarchy, Chen2021f, Wang2022a}  view hierarchy as a directed acyclic graph and model label hierarchy based on graph encoders.
However, the input of graph encoders is static, considering that all HTC text samples share the same hierarchical structure, which leads graph encoders to model the same graph redundantly.
To solve this problem, \citeauthor{Wang2022a}~(\citeyear{Wang2022a}) directly discards the graph encoder during prediction, but this method still suffers from the same problem during training.
Moreover, since the target labels corresponding to each text sample could be either a single-path or a multi-path in HTC~\cite{zhou2020hierarchy}, recent methods only consider the graph of the entire label hierarchy and ignore the subgraph corresponding to each text sample.
This subgraph can contain structured label co-occurrence information.
For instance, a news report about France travel is labeled ``European'' under the parent label ``World'' and ``France'' under a different parent label ``Travel Destinations''.
There is a strong correlation between the labels ``France'' and ``European''. But these labels are far apart on the graph, making it difficult for graph encoders to model this relationship.

Under such observation, we divide the label hierarchy into global and local hierarchies to take full advantage of hierarchical information in HTC.
We define global hierarchy as the whole hierarchical structure, referred to as hierarchical information in previous methods.
Then we define local hierarchy as a structured label hierarchy corresponding to each text sample, which is the subgraph of global hierarchy.
Moreover, global hierarchy is static and irrelevant to text samples, while local hierarchy is dynamic and relevant to text samples.
Considering the characteristics of two hierarchies,
our method models them separately to avoid redundantly modeling static global hierarchy and fully exploit hierarchical information with dynamic local hierarchy.

To model semantic information along with hierarchical information, \citeauthor{zhou2020hierarchy}~(\citeyear{zhou2020hierarchy}) proposes hierarchy-aware multi-label attention. \citeauthor{Chen2021f}~(\citeyear{Chen2021f}) reformulates it as a matching problem by encouraging the text representation to be similar to its label representation.
Although,  these methods can improve the performance of text encoders by injecting label hierarchy with the graph encoder
,
the improvement on pretrained language model BERT~\cite{devlin2018bert} is limited~\cite{Wang2022a}.
Compared to previous text encoders such as CNN or RNN, BERT has large-scale parameters and prior language knowledge.
It enables BERT to roughly grasp hierarchical information with multi-label text classification.
Therefore, HGCLR~\cite{Wang2022a} is proposed to improve the BERT performance on HTC, and the hierarchy is embedded into BERT based on contrastive learning during training. For prediction, HGCLR directly uses BERT as a multi-label classifier.
Specifically, the hierarchy in HGCLR is represented by positive samples in contrastive learning, which is implemented by scaling the BERT token embeddings based on a graph encoder.
However, representing hierarchy by simply scaling token embeddings is inefficient, which may also lead to a gap between training and prediction.

To efficiently exploit BERT in HTC, we leverage the prior knowledge of BERT by transforming both global and local hierarchy modeling as mask prediction tasks.
Moreover, we discard the auxiliary graph encoder and utilize BERT to model hierarchical information to avoid the intentional fusion of BERT and graph encoder.
For global hierarchy, we propose a label mask prediction task to recover masked labels based on the label relationship in global hierarchy.
Since global hierarchy is irrelevant to text samples, we only fine-tune label embeddings and keep BERT frozen. For local hierarchy, we combine text samples and labels as the input of BERT to directly fuse semantic and hierarchical information according to the attention mechanism in BERT.

In summary, the contributions  of this paper are following:
\begin{itemize}
    \item We propose HBGL to take full advantage of BERT in HTC. HBGL does not require auxiliary modules like graph encoders to model hierarchical information, which avoids the intentional fusion of semantic and hierarchical modules.
    \item We propose corresponding methods to model the global and local hierarchies based on their characteristics in order to further exploit the information of the hierarchy.
    \item
      Experiments show that the proposed model achieves significant improvements on three datasets. Our code will be public to ensure reproducibility.
\end{itemize}

\begin{figure*}[t]
\centering
\includegraphics[width=\linewidth]{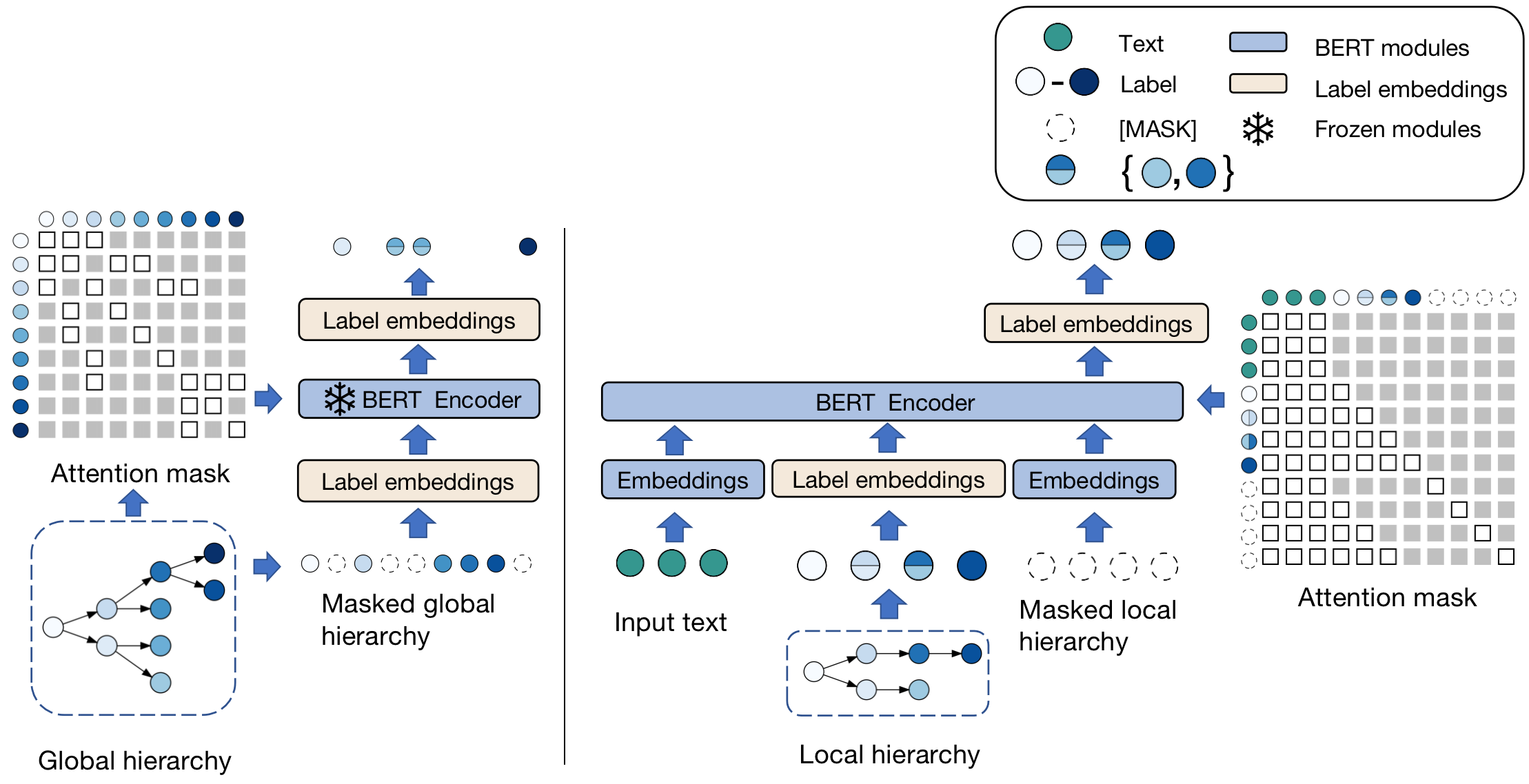}
\caption{The overall framework of our model under the label hierarchy with four maximum levels.
The left part is the global hierarchy-aware label embeddings module. The right part is the local hierarchy-aware text encoder module.
We use different colors to identify labels, and the darker color indicates the lower level.
Gray squares in attention masks indicate that tokens are prevented from attending,
while white squares indicate that attention between tokens is allowed.
The label embeddings share the same weight in both modules, which is initialized by the global hierarchy-aware label embeddings module.
Note the special tokens like \texttt{[CLS]} or \texttt{[SEP]}, position and segment tokens of BERT are ignored for simplicity, which we will discuss in methodology.
(Best view in color.)
}
\label{fig:fig1}
\end{figure*}

\section{Related Work}
Hierarchical text classification (HTC) is a special multi-label text classification problem that requires constructing one or more paths from the taxonomic hierarchy in a top-down manner~\cite{sun2001hierarchical}.
Compared to multi-label text classification, HTC focuses on leveraging hierarchical information to achieve better results.
There are two groups of existing HTC methods based on treating the label hierarchy: local and global approaches.

The local approaches leverage hierarchical information by constructing one or more classifiers at each level or each node in hierarchy. Generally speaking, a text sample will be classified top-down according to its hierarchy.
\citeauthor{shimura-etal-2018-hft}~(\citeyear{shimura-etal-2018-hft}) applies a CNN with a fine-tuning technique to utilize the data in the upper levels.
\citeauthor{banerjee-etal-2019-hierarchical}~(\citeyear{banerjee-etal-2019-hierarchical}) initials parameters of the child classifier by the fine-tuned parent classifiers.

The global approaches leverage hierarchical information by directly treating HTC as multi-label text classification with hierarchy information as input.
Many methods like recursive regularization~\cite{10.1145/2487575.2487644}, reinforcement learning~\cite{mao-etal-2019-hierarchical}, capsule network~\cite{8933476}, and meta-learning~\cite{wu-etal-2019-learning} has been proposed to capture hierarchical information.
To better represent hierarchical information,
\citeauthor{zhou2020hierarchy}~(\citeyear{zhou2020hierarchy}) formulates the hierarchy as a directed graph and introduces hierarchy-aware structure encoders.
\citeauthor{Chen2021f}~(\citeyear{Chen2021f}) formulates the text-label semantics relationship as a semantic matching problem.

With the development of Pretrained Language Model (PLM),
PLM outperforms previous methods even without using hierarchical information.
Compared to text encoders like RNN or CNN, PLM is strong enough to learn hierarchical information without hierarchy-aware structure encoders.
Under this observation,
\citeauthor{Wang2022a}~(\citeyear{Wang2022a}) proposes HGCLR to embed the hierarchical information into the PLM directly.
However, HGCLR still requires the hierarchy-aware structure encoder like Graphormer~\cite{ying2021transformers} to incorporate hierarchical information during training.


\section{Problem Definition}

Given a training set \(\{(\mathbf{x}_i, \mathbf{y}_i)\}^{N}_{i=1}\) where $\mathbf{x}_i$ is raw text, and $\mathbf{y}_i \in \{0,1\}^L$ is the label of $\mathbf{x}_i$ represented by $L$ dimensional multi-hot vector.
The goal of Hierarchical Text Classification (HTC) is to predict a subset labels for $\mathbf{x}_i$ with the help of hierarchical information, which can be organized as a Directed Acyclic Graph (DAG) $G=(V,\overrightarrow{E}, \overleftarrow{E})$, where $V=\{v_1,\ldots,v_L\}$ is the set of label nodes.
$\overrightarrow{E}=\{(v_i,v_j)|v_j \in child(v_i)\}$ is the top-down hierarchy path and
$\overleftarrow{E}=\{(v_j,v_i)|v_j \in child(v_i)\}$ is the bottom-up hierarchy path.
Although labels in $\mathbf{y}_i$ follow the labels hierarchy, HTC could be either a single-path or a multi-path problem~\cite{zhou2020hierarchy}.

\section{Methodology}
In this section, we provide the technical details of the proposed HBGL.
Figure \ref{fig:fig1} shows the overall framework of the model.
The left part corresponds to global hierarchy-aware label embeddings, while the right part corresponds to local hierarchy-aware text encoder.
We first inject global hierarchy into label embeddings in global hierarchy-aware label embeddings. Then we leverage these label embeddings with our local hierarchy-aware text encoder.

\subsection{Global Hierarchy-aware Label Embeddings}

Global hierarchy-aware label embeddings aims to initialize the label embeddings based on the label semantics and hierarchy in HTC.
It allows BERT to directly exploit the global hierarchy without auxiliary label encoders.
Contrary to previous methods~\cite{zhou2020hierarchy, Chen2021f, Wang2022a},
we implement BERT as a graph encoder to initialize the label embeddings via gradients descent and adapt the global hierarchy to label embeddings by formulating it as mask prediction.
Global hierarchy-aware label embeddings leverages the large-scale pretraining knowledge of BERT to generate better hierarchy-aware and semantic-aware label embeddings.

Following \citeauthor{Wang2022a}~(\citeyear{Wang2022a}), we first initialize label embeddings $\mathbf{\hat{Y}} = [\mathbf{\hat{y}}_1,\ldots ,\mathbf{\hat{y}}_L]  \in \mathbb{R}^{L \times d}$ with the averaging BERT token embeddings of label name, where $d$ is the hidden size of BERT and $\hat{\mathbf{y}}_i$ corresponds to label node $v_i$ in $G$.
Since $\mathbf{\hat{Y}}$ is initialized with the BERT token embedding, it takes advantage of the prior knowledge of BERT to merge label semantic and hierarchical information.
Specifically, the input embeddings $\mathbf{e}  \in \mathbb{R}^{L \times d}$ of BERT encoder is defined as:
\begin{equation}
\label{eq:hle-emb}
\mathbf{e}_{i} = {\mathbf{\hat{y}}}_{i} + \mathbf{t}_1 + \mathbf{p}_{{\rm HLevel}(v_{i})} \\
\end{equation}
where $\mathbf{p}  \in \mathbb{R}^{512 \times d}$ and $\mathbf{t} \in \mathbb{R}^{2 \times d}$ are the position embeddings and segment embeddings in BERT~\cite{devlin2018bert}.
To exploit the position embeddings $\mathbf{p}$, 
we use the hierarchy level ${\rm HLevel}(v_i)$ as the position id for label $v_i$.
Low position ids represent coarse-grained labels, while high position ids represent fine-grained labels.
To exploit the segment embeddings $\mathbf{t}$, we use segment id $1$ to represent labels, which makes BERT easy to distinguish labels and text in HTC.

To feed BERT with the label graph, we add attention mask $\mathbf{A}\in \{0,1\}^{L \times L}$ in each self-attention layers.
Formally, $\mathbf{A}$ is defined as:

\begin{equation}
\label{eq:hle-attn}
\mathbf{A}_{ij} = \begin{cases}
0, &\text {if} (v_i, v_j) \in \overrightarrow{E} \cup \overleftarrow{E} \text{ or } i = j \\
1, & \text{otherwise} \end{cases}  
\end{equation}
where
$\overrightarrow{E}$ is the top-down hierarchy path and
$\overleftarrow{E}$ is the bottom-up hierarchy path of DAG $G$.
We allow one label can attend its parent and child labels.
For example,
we show the attention mask for the four-level hierarchy in Figure \ref{fig:fig1}.

Based on input embeddings $\mathbf{E}$ and attention mask $\mathbf{A}$, we can use mask LM task~\cite{devlin2018bert} to inject hierarchy into $\mathbf{\hat{Y}}$.
However, if we directly follow mask LM task in BERT,
it will cause BERT unable to distinguish between masked leaf labels under the same parent label.
For example, two leaf labels ``Baseball'' and ``Football'' under the same parent label ``Sport''.
The model will output the same result if both leaf labels are masked.
Since both labels have the same position and segment embeddings, and only attend to ``Sport'' label in $\mathbf{A}$.
To solve this problem, we treat the masked label prediction task as the multi-label classification, which requires a masked leaf label to predict itself and other masked sibling leaf labels according to $G$.

Formally, we first random mask several labels by replacing $\mathbf{\hat{y}}_i$ with mask token embedding in Eq. \ref{eq:hle-emb} to get masked input embeddings $\mathbf{e}^{\prime}$.
Second, we calculate the hidden state representation $\mathbf{h}\in \mathbb{R}^{L \times d}$ and scores of each label $\mathbf{s} \in \mathbb{R}^{L \times L}$ as following:
\begin{equation}
\label{eq:hle-bert}
\begin{split}
\mathbf{h}&=\text{BERTEncoder}(\mathbf{e}^{\prime}, \mathbf{A})\\
\mathbf{s} &= \text{sigmoid}(\mathbf{h}\mathbf{\hat{Y}}^T)
\end{split}
\end{equation}
Where \text{BERTEncoder} is the encoder part of BERT and $\mathbf{A}$ is applied to each layer of \text{BERTEncoder}.
Finally, The problem of injecting hierarchy into $\hat{\mathbf{Y}}$ can be reformulated as solving the following optimization problem:
\begin{equation}
 \label{eq:hle-loss}
\begin{aligned}
  &\min _{\hat{\mathbf{Y}}}  \mathcal{L}_\text{global}=\\
  &-\sum_{i}^{i\in V_m}\sum_{j=1}^{L}\left[\bar{\mathbf{y}}_{ij} \log \left(\mathbf{s}_{ij}\right)+\left(1-\bar{\mathbf{y}}_{ij}\right) \log \left(1-\mathbf{s}_{ij}\right)\right]
\end{aligned}
\end{equation}
Where $V_m$ is the masked labels set and $\bar{\mathbf{y}}_{ij}$ is the target for $\mathbf{s}_{ij}$. We set $\bar{\mathbf{y}}_{ij} = 1 $, when $i = j$ or the label $i$ and $j$ are masked sibling leaf nodes in $G$.
To avoid model overfitting on static graph $G$, we keep all parameters of  BERT  frozen and only fine-tune label embeddings $\hat{\mathbf{Y}}$ in Eq. \ref{eq:hle-loss}.
Moreover, we gradually increase the label mask ratio during training.

The whole procedure of global hierarchy-aware label embeddings is shown in Algorithm~\ref{alg:hle}.

\begin{algorithm}[h!]
	\caption{\small{Global Hierarchy-aware Label Embeddings}}
	\label{alg:hle}
	{\bf input:} Label hierarchy $G$ and label names \\
	{\bf output:} Label embeddings $\hat{\mathbf{Y}}$.\\
	{\bf initialize:} $\hat{\mathbf{Y}}$ using averaging BERT token embeddings of each label name.

	\begin{algorithmic}[1]
    \STATE Set mask ratio $r_m$;
    \STATE Set mask ratio upper bound $r_{M}$;
    \STATE Set learning rate $lr$, batch size $bsz$ and training steps $T_{train}$;
    \STATE Get attention mask $\mathbf{A}$ according to Eq. \ref{eq:hle-attn};
		\FOR {$t = 1, ..., T_{train} $}
    \STATE Get input embeddings $e$ according to Eq. \ref{eq:hle-emb};
		\FOR {$b = 1, ..., bsz $}
    \STATE Mask $e$ with mask ratio $r_m^t$ to get $\mathbf{e^{\prime}}$;
    \STATE Get $\mathbf{h}$ and $\mathbf{s}$ according to Eq. \ref{eq:hle-bert};
    \STATE Get $\bar{\mathbf{y}}$ based on $\mathbf{e^{\prime}}$ and $G$,
    $\bar{\mathbf{y}}_{ij}=1$, when $j = i$ or the label $i$ and $j$ are masked sibling leaf nodes;
    \STATE Compute loss $\mathcal{L}$ in Eq. \ref{eq:hle-loss};
    \STATE Backward and compute the gradients $\frac{\partial \mathcal{L}}{\partial \mathbf{\hat{Y}}^{tb}}$;
    \STATE Accumulate gradients $\frac{\partial \mathcal{L}}{\partial \hat{\mathbf{Y}}^{tb}}$ to $\frac{\partial \mathcal{L}}{\partial \hat{\mathbf{Y}}^{t}}$
		\ENDFOR
    \STATE $r_m^{t+1} = r_m^t + \frac{r_{M} - r_m}{T_{train}}$
		\STATE $ \hat{\mathbf{Y}}^{t+1} = \text{UpdateParameter}(\hat{\mathbf{Y}}^t, \frac{\partial \mathcal{L}}{\partial \hat{\mathbf{Y}}^t},lr)$;
		\ENDFOR
	\end{algorithmic}
\end{algorithm}

\subsection{Local Hierarchy-aware Text Encoder}

Local hierarchy is the structured label hierarchy corresponding to each text sample, which is ignored in previous methods~\cite{zhou2020hierarchy, Chen2021f, Wang2022a}.
In contrast to global hierarchy,
local hierarchy is dynamic and related to semantic information.
To leverage local hierarchy in HTC, we need to examine several issues before introducing our methods.
First, although local hierarchy contains the hierarchical information related to target labels, this leads to label leakage during training.
Second, we should pay attention to the gap between training and prediction, since local hierarchy is only available during training.
To this end, we propose local hierarchy-aware text encoder to exploit local hierarchy while avoiding the above issues.

\subsubsection{Local Hierarchy Representation}
We first discuss the representation of local hierarchy in BERT before introducing local hierarchy-aware text encoder.
Following the method in global hierarchy-aware label embeddings,
we can represent local hierarchy as the subgraph of global hierarchy according to attention mechanism of BERT.
However, it is hard for BERT to combine the input of label graph and text sample, while avoiding label leakage and the gap between training and prediction.
Therefore, we propose another method to efficiently represent local hierarchy, which allows BERT to combine local hierarchies with text samples easily.
Since the local hierarchy is single-path or multi-path in the global hierarchy, in the single-path case we can simply treat it as the sequence.
If we can also transform the multi-path case into the sequence, we can represent local hierarchy as the sequence, making it easy to model with BERT.
Under this observation, we use the following method to transform the multi-path local hierarchy into the sequence:

\begin{equation}
\label{eq:lr-rep}
\begin{split}
\mathbf{u}_h &= \sum_{j}^{j\in y^h}{\hat{\mathbf{y}}_j}\\
\mathbf{u} &= [\mathbf{u}_1,\ldots,\mathbf{u}_D]
\end{split}
\end{equation}
Where $\mathbf{u}_h \in \mathbb{R}^{d}$ is the $h$th level of hierarchy, $y^h$ is the target labels in $h$th level, $\hat{\mathbf{y}}_j$ is the global hierarchy-aware label embeddings, $D$ is the maximum level of hierarchy and $\mathbf{u}$ is local hierarchy sequence.

For example, consider a multi-path local hierarchy with four labels: $1a$, $1b$, $2a$ and $2b$, where $2a$ and $2b$ are the child labels of $1a$ and $1b$, respectively.
According to Eq. \ref{eq:lr-rep}, we can get $\mathbf{u}_1 = \mathbf{\hat{y}}_{1a} + \mathbf{\hat{y}}_{1b}$ and $\mathbf{u}_2 = \mathbf{\hat{y}}_{2a} + \mathbf{\hat{y}}_{2b}$. The attention score $\alpha_{2 1}$  in BERT can be calculated as:
\begin{equation}
\label{eq:lr-e}
\begin{split}
 \alpha_{21} =& \mathbf{u}_{2}W^Q(\mathbf{u}_{1}W^K)^T\\
 = & \mathbf{\hat{y}}_{2a}W^Q(\mathbf{\hat{y}}_{1a}W^K)^T+\mathbf{\hat{y}}_{2b}W^Q(\mathbf{\hat{y}}_{1b}W^K)^T \\
  & +\mathbf{\hat{y}}_{2a}W^Q(\mathbf{\hat{y}}_{1b}W^K)^T+\mathbf{\hat{y}}_{2b}W^Q(\mathbf{\hat{y}}_{1a}W^K)^T
\end{split}
\end{equation}
Where $W^Q, W^K \in \mathbb{R}^{d \times d_z}$ are parameter matrices in BERT.
As shown in Eq. \ref{eq:lr-e},
$\alpha_{21}$ contains  $\mathbf{\hat{y}}_{2a}W^Q(\mathbf{\hat{y}}_{1a}W^K)^T$ and $\mathbf{\hat{y}}_{2b}W^Q(\mathbf{\hat{y}}_{1b}W^K)^T$ to represent the local hierarchy graph,
while it also contains $\mathbf{\hat{y}}_{2a}W^Q(\mathbf{\hat{y}}_{1b}W^K)^T$ and $\mathbf{\hat{y}}_{2b}W^Q(\mathbf{\hat{y}}_{1a}W^K)^T$.
Since we have injected global hierarchy into the label embeddings: $\mathbf{\hat{y}}_{1a}$, $\mathbf{\hat{y}}_{1b}$, $\mathbf{\hat{y}}_{2a}$ and $\mathbf{\hat{y}}_{2b}$ according to Algorithm~\ref{alg:hle}, which leverages the first part in $\alpha_{21}$ to predict masked labels,
it allows $\alpha_{21}$ to be able to hold hierarchical information in local hierarchy.
In addition, the second part of $\alpha_{21}$ is also relevant for modeling the local hierarchy, as the labels in it correspond to the same text sample.

\subsubsection{Fusing Local Hierarchy into Text Encoder}
To further exploit local hierarchy, while avoiding label leakage and the gap between training and prediction, it is hard to implement BERT directly as multi-label classifier.
Inspired by s2s-ft~\cite{bao2021s2s}, which adopts PLM like BERT for sequence-to-sequence learning, we propose a novel method by adopting BERT to generate the local hierarchy sequence.
Note that since elements in the local hierarchy sequence may contain multiple labels, we cannot use sequence-to-sequence methods directly.

In order to fuse local hierarchy and text in sequence-to-sequence fashion,
our model aims to generate the local hierarchy sequence $\textbf{u}$:
\begin{equation}
  \label{eq:lr-p}
p(\textbf{u} \mid \textbf{x})=\prod_{h=1}^{D} p\left(\textbf{u}_{h} \mid \textbf{u}_{<h}, \textbf{x}\right)
\end{equation}
where  $\textbf{u}_{<h} = \mathbf{u}_1,\ldots,\mathbf{u}_{h-1}$ and $\textbf{x}$ is the input text corresponding to $\textbf{u}$.
There are several advantages to model HTC as Eq. \ref{eq:lr-p}:
First, the structure of local hierarchy can be included. Since $\textbf{u}_{h}$ represents the labels corresponding to the $h$th level of hierarchy, it only depends on the labels above $h$th level, which can been represented by $p\left(\textbf{u}_{h} \mid \textbf{u}_{<h}, x\right)$.
Second, we can leverage the teacher forcing to fuse local hierarchy and text while avoiding label leakage during training.

Specifically, the input of BERT is composed of three parts:
input text, local hierarchy and masked local hierarchy during training, as shown in Figure \ref{fig:fig1}.
The end-of-sequence token \texttt{[SEP]} is used to divide these three parts.
Based on s2s-ft,
we implement similar attention mask provided in Figure \ref{fig:fig1}.
The attention mask prevents the input text from attending to the local hierarchy and masked local hierarchy, which guarantees that labels do not influence the input text tokens.
The attention mask also ensures the top-down manner in the local hierarchy, which allows the label to attend to the upper level labels in the hierarchy.
For the attention mask between local hierarchy and masked local hierarchy, it allows the masked labels to be predicted based on upper level target labels, following the teacher forcing manner.
We use 0 and 1 to distinguish text and label for segment ids in BERT.
For position ids in BERT, we set the same position ids in the local hierarchy and mask local hierarchy by accumulating them based on text location ids.
By feeding BERT with the above inputs,
 we use a binary cross-entropy loss function to predict labels in each level separately. The optimization problem is as follows:
\begin{equation}
\begin{aligned}
  &\min _{\Theta,\hat{\mathbf{Y}}}  \mathcal{L}_\text{local}=\\
  &-\!\sum_{i=1}^{N}\!\sum_{h=1}^{D}\!\sum_{j}^{j\in V_h}\!\left[\mathbf{y}_{ij} \log\! \left({\mathbf{s}}^t_{ihj}\!\right)\!+\!\left(1\!-\!\mathbf{y}_{ij}\!\right) \log\! \left(1\!-\!{\mathbf{s}}^t_{ihj}\!\right)\right]
\end{aligned}
\end{equation}
where $\Theta$ are parameters of BERT, $V_h = \{ j\ |\ {\rm HLevel}(v_j) = h  \}$ is the labels in $h$th level, $\mathbf{y}_{ij}$ is the $j$th label corresponding to text $\mathbf{x}_i$ and ${\mathbf{s}}^t_{ihj}$ is the score of $j$th label, which is calculated as following:
\begin{equation}
  {\mathbf{s}}^t_{ih} = \text{sigmoid}({\mathbf{h}}^t_{ih}\mathbf{\hat{Y}}^T)
\end{equation}
where ${\mathbf{h}}^t_{ih} \in \mathbb{R}^d$ is the hidden state representation of $h$th masked local hierarchy corresponding to $\mathbf{x}_i$ and ${\mathbf{s}}^t_{ihj}$ is the $j$th element of ${\mathbf{s}}^t_{ih} \in \mathbb{R}^L$.

For prediction, we utilize the local hierarchy during the training stage to make BERT separately predict labels at each level in an autoregressive manner.
The scores $\mathbf{s}^p_i$  for $i$th level labels are computed as following:
\begin{equation}
\begin{split}
\mathbf{h}^p_h&=\text{BERTEncoder}([\mathbf{e}_\text{text};\mathbf{u}^p_{<h};\mathbf{e}_\text{mask} ], \mathbf{A}^{p_h})\\
\mathbf{s}^p_h &=\text{sigmoid}(\mathbf{h}^p_{h}\mathbf{\hat{Y}}^T)\\
\end{split}
\end{equation}
where  $\mathbf{h}^p_h \in \mathbb{R}^d$ is the hidden state representation of $h$th level,
$\mathbf{e}_\text{text}$ and $\mathbf{e}_\text{mask}$ are text embeddings and mask token embedding, $\mathbf{A}^{p_h}$ is the attention mask for $h$th level, which is a submatrix of the attention mask between text and local hierarchy in training, and $\mathbf{u}^p_{<h} =  \mathbf{u}^p_{1} \cdots \mathbf{u}^p_{h-1} $ is:
\begin{equation}
\begin{gathered}
\mathbf{u}^p_h = \begin{cases}
  \mathbf{e}_\text{sep}, &\text {if} \sum_{j=0}^L \mathbbm{1}(\mathbf{s}^p_{hj}) = 0\\
  \sum_{j=0}^L \mathbbm{1}(\mathbf{s}^p_{hj}) \mathbf{\hat{y}}_j^T, & \text{otherwise} \end{cases}  \\
s . t \mathbbm{1}\left(\mathbf{s}^p_{hj}\right)= \begin{cases}1, & \text { if } \mathbf{s}^p_{hj} > 0.5  \\ 0, & \text { otherwise }\end{cases}
\end{gathered}
\end{equation}
where $\mathbbm{1}\left(\mathbf{s}^p_{hj}\right)$ represents the predicted label corresponding to $\mathbf{s}^p_{hj}$.
Specifically,
we sum $h$th level predicted label embeddings to generate $\mathbf{u}^p_h$ and replace $\mathbf{u}^p_h$ with  $\mathbf{e}_\text{sep}$ \texttt{[SEP]} token embedding when these is no predicted label in $h$th level.

Finally, the predicted labels set $y^p$ is:
\begin{equation}
  y^p = \{v_j |  \mathbf{s}^p_{hj} > 0.5 \text{ and } h = {\rm HLevel}(v_j) \}
\end{equation}

\section{Experiments}

\paragraph{Datasets and Evaluation Metrics}
We select three widely-used HTC benchmark datasets in our experiments. They are: Web-of-Science (WOS)~\cite{kowsari2017hdltex}, NYTimes (NYT)~\cite{shimura-etal-2018-hft}, and RCV1-V2~\cite{10.5555/1005332.1005345}. The detailed information of each dataset is shown in Table~\ref{tab:data_statistic}. We follow the data processing of previous works~\cite{zhou2020hierarchy, Wang2022a} and use the same evaluation metrics to measure the experimental results: Macro-F1 and Micro-F1.

\begin{table}[h]
\resizebox{\linewidth}{!}{
\begin{tabular}{c|cccccc}
\toprule
Dataset & $L$   & $D$ &   Avg($L_i$)   & Train  & Dev   & Test    \\ \midrule
WOS     & 141 & 2     & 2.0  & 30,070 & 7,518 & 9,397   \\
NYT     & 166 & 8     & 7.6  & 23,345 & 5,834 & 7,292   \\
RCV1-V2    & 103 & 4     & 3.24 & 20,833 & 2,316 & 781,265 \\ \bottomrule
\end{tabular}
}
\caption{Data Statistics. $L$ is the number of classes. $D$ is the maximum level of hierarchy. Avg($|L_i|$) is the average number of classes per sample.}
\label{tab:data_statistic}
\end{table}

\begin{table*}[t]
\resizebox{\textwidth}{!}{%
\begin{tabular}{@{}ccccccc@{}}
\toprule
\multirow{2}{*}{Model}              & \multicolumn{2}{c}{WOS} & \multicolumn{2}{c}{NYT} & \multicolumn{2}{c}{RCV1-V2} \\ \cmidrule(l){2-7}
                                    & Micro-F1                & Macro-F1                & Micro-F1         & Macro-F1         & Micro-F1         & Macro-F1     \\ \midrule
\multicolumn{7}{c}{\textbf{Hierarchy-Aware Models}}                                                                        \\ \midrule
TextRCNN \cite{zhou2020hierarchy}   & 83.55                   & 76.99                   & 70.83            & 56.18            & 81.57            & 59.25        \\
HiAGM \cite{zhou2020hierarchy}      & 85.82                   & 80.28                   & 74.97            & 60.83            & 83.96            & 63.35        \\
HTCInfoMax \cite{deng-etal-2021-htcinfomax}& 85.58                   & 80.05                   & -                & -                & 83.51            & 62.71        \\
HiMatch \cite{Chen2021f}        & 86.20                   & 80.53                   & -                & -                & 84.73            & 64.11        \\ \midrule
\multicolumn{7}{c}{\textbf{Pretrained Language Models}}                                                                    \\ \midrule
  BERT$^\dagger$                             & 85.63                   & 79.07                   & 78.24            & 65.62            & 85.65            & 67.02        \\
  BERT+HiAGM$^\dagger$                        & 86.04                   &  80.19                  & 78.64            & 66.76            &  85.58           & 67.93        \\
  BERT+HTCInfoMax$^\dagger$                  &  86.30                  &  79.97                  & 78.75            & 67.31            &  85.53           &  67.09      \\
BERT+HiMatch \cite{Chen2021f}   & 86.70                   & 81.06                   & -                & -                & 86.33            & 68.66        \\
HGCLR~\cite{Wang2022a}                               &  87.11                  &   81.20                 &   78.86          &   67.96          &   86.49          &  68.31      \\
HBGL &  \textbf{87.36}         &   \textbf{82.00}        &   \textbf{80.47} &   \textbf{70.19} &   \textbf{87.23} &  \textbf{71.07}      \\
\bottomrule
\end{tabular}%
}
\caption{Experimental results of our proposed model on several datasets. $\dagger$: results from \cite{Wang2022a}. }
\label{tab:2}
\end{table*}
\subsection{implement Details}
Following HGCLR~\cite{Wang2022a}, we use \texttt{bert-base-uncased} as both text and graph encoders.
The graph structure input of BERT is implemented based on the attention mask of huggingface transformers~\cite{wolf-etal-2020-transformers}.
We introduce the implementation details of the global hierarchy-aware label embeddings and the local hierarchy-aware text encoder, respectively.

For global hierarchy-aware label embeddings, we first initialize the label embeddings by averaging their label name token embeddings in BERT.
To embed global hierarchy into label embeddings, we train initialized label embeddings with frozen \texttt{bert-base-uncased} according to Algorithm~\ref{alg:hle}.
The initial mask ratio and mask ratio upper bound are 0.15 and 0.45, respectively.
Considering the different maximum levels of hierarchy and the number of labels in each dataset, we grid search learning rates of label embeddings among \{1e-3, 1e-4\} and the training steps among \{300, 500, 1000\}.

For local hierarchy-aware text encoder,
we follow the settings of HGCLR.
The batch size is set to 12. The optimizer is Adam with a learning rate of 3e-5.
We use global hierarchy-aware label embeddings to initialize the label embeddings.
And the input label embeddings share weights with the label embeddings in the classification head.
The input label length of each dataset is set to the depth in Table~\ref{tab:data_statistic}.
Compared to the 512 maximum input tokens in HGCLR, we use smaller input tokens to achieve similar prediction time performance.
According to the maximum hierarchy level, the maximum input tokens in WOS, NYT, and RCV1-V2 are 509, 472, and 492, respectively.
Moreover, we cache the previous attention query and key values to make the prediction more efficient.

\subsection{Baselines}
We compared the state-of-the-art and most enlightening methods including HiAGM~\cite{zhou2020hierarchy}, HTCInfoMax~\cite{deng-etal-2021-htcinfomax}, HiMatch~\cite{Chen2021f}, and HGCLR~\cite{Wang2022a}.
HiAGM, HTCInfoMax, and HiMatch use different fusion strategies to mix text-hierarchy representation.
Specifically,
HiAGM proposes hierarchy-aware multi-label attention to get the tex-hierarchy representation.
HTCInfoMax introduces information maximization to model the interaction between text and hierarchy.
HiMatch reformulates it as a matching problem by encouraging the text representation be similar to its hierarchical label representation.
Contrary to the above methods,
HGCLR achieves state-of-the-art results by directly incorporating hierarchy into BERT based on contrastive learning.

\subsection{Experimental Results}
Table~\ref{tab:2} shows Micro-F1 and Macro-F1 on three datasets.
Our method significantly outperforms all methods by further exploiting the hierarchical information of HTC and the prior knowledge of BERT.

Compared to BERT, we show proposed HBGL can significantly leverage the hierarchical information by achieving 2.99\%, 4.59\% and 4.05\% improvement of Macro-F1 on WOS, NYT, and RCV1-V2. By the way, our method shows better performance on the dataset with a complex hierarchy.
Our method can achieve 4.59\% improvement of Macro-F1 on NYT with the largest label depth and number of labels in the three datasets.

HBGL also shows impressive performance compared to BERT based HTC methods.
Current state-of-the-art methods like HGCLR, which relies on contrastive learning to embed the hierarchy into BERT, has negligible improvement over previous methods such as HiMatch.
Furthermore, although different methods of incorporating semantic and hierarchical information are used, these methods have similar performances on three datasets, which shows a common limitation in previous methods: merging BERT and the graph encoder regardless of local hierarchy and prior knowledge of BERT.
By overcoming this limitation, our method observes 2.23\% and 2.76\% boost on Macro-F1 on NYT and RCV1-V2 compared to HGCLR.
For WOS,
it is the simplest dataset among the above datasets with two-level label hierarchy, and the labels for each document are single-path in the hierarchy, the impact of leveraging hierarchy is smaller than the other two datasets.
However, our methods still achieve reasonable improvement compared to HGCLR.

\subsection{Effect of Global Hierarchy-aware Label Embeddings}

To examine the effectiveness of global hierarchy-aware label embeddings, we compare four label embeddings methods:
global hierarchy-aware label embeddings (global hierarchy BERT),
global hierarchy-aware label embeddings, where BERT is replaced with GAT (global hierarchy GAT),
label embeddings initialized by averaging BERT token embeddings in each label name (label name) and
randomly initialize label embeddings (random).
For global hierarchy GAT and BERT, we use label names as the initialized label embeddings.
For global hierarchy GAT,  we get the best results by grid searching learning rate, training step and whether to freeze GAT.

As shown in Table \ref{tab:3}, our method outperforms the other three label embeddings methods.
Although all methods can leverage hierarchical information by the local hierarchy-aware text encoder, the remaining methods still achieve poorer performance than global hierarchy BERT, which shows the importance of global hierarchy.
For global hierarchy GAT, GAT cannot leverage global hierarchy by predicting masked labels like BERT.

\begin{table}[h]
\resizebox{\linewidth}{!}{
\begin{tabular}{lcccc}
\toprule
\multirow{2}{*}{Label  Embeddings} & \multicolumn{2}{c}{NYT} & \multicolumn{2}{c}{RCV1-V2} \\
                                   & Micro-F1                & Macro-F1       & Micro-F1        & Macro-F1     \\ \midrule
Random                             & 79.18                   & 67.92          &  86.98          & 70.15       \\
Label name                         & 80.26                   & 69.64          &  87.20          & 70.24\\
Global hierarchy GAT               & 80.15                   & 69.59          & 86.69           & 70.23 \\
Global hierarchy BERT                   & \textbf{80.47}          & \textbf{70.19} &  \textbf{87.23} & \textbf{71.07}\\ \bottomrule
\end{tabular}
}
\caption{Impact of different label embeddings on NYT and RCV1-V2.}
\label{tab:3}
\end{table}

\subsection{Effect of Local Hierarchy-aware Text Encoder}

We also analyze the importance of local hierarchy-aware text encoder by comparing it with two methods:
multi-label and seq2seq.
For multi-label, we fine-tune BERT as the multi-label classifier.
For seq2seq, we fine-tune BERT as the seq2seq model following s2s-ft, where target labels are sorted according to their levels in the global hierarchy.
Local hierarchy-aware text encoder achieves the best performance in Table \ref{tab:4}.

\begin{table}[h]
\resizebox{\linewidth}{!}{
\begin{tabular}{lcccc}
\toprule
\multirow{2}{*}{Fine tuning} & \multicolumn{2}{c}{NYT} & \multicolumn{2}{c}{RCV1-V2} \\
                             & Micro-F1                & Macro-F1       & Micro-F1        & Macro-F1     \\ \midrule
Mulit-label                  & 78.16                   & 67.05          &  85.96          &  68.03       \\
Seq2seq                      & 79.22                   & 67.82          &  86.22          &  67.97       \\
Local hierarchy              & \textbf{80.47}          & \textbf{70.19} &  \textbf{87.23} & \textbf{71.07}\\ \bottomrule
\end{tabular}
}
\caption{Impact of different fine tuning methods on NYT and RCV1-V2.}
\label{tab:4}
\end{table}

\section{Conclusion}
In this paper, we proposed a BERT-based HTC framework HBGL.
HBGL avoids the intentional fusion of semantic and hierarchical modules by utilizing BERT to model both semantic and hierarchical information.
Moreover, HBGL takes full advantage of hierarchical information by modeling global and local hierarchies, respectively.
Considering that global hierarchy is static and irrelevant to text samples,  we propose global hierarchy-aware label embeddings to inject global hierarchy into label embeddings directly.
Considering that local hierarchy is dynamic and relevant to text samples, we propose local hierarchy-aware text encoder to deeply combine semantic and hierarchical information according to the attention mechanism in BERT.
Compared to existing methods, HBGL achieves significant improvements on all three datasets, while only parameters corresponding to label embeddings are required except BERT.

\section{Limitation}
While HBGL exploits global and local hierarchies and achieves improvements on three HTC datasets,
one limitation is that HBGL requires additional iterations to predict labels.
HBGL needs to predict upper level labels before predicting current level labels.
To alleviate this limitation,  we cached the BERT attention query and key values from previous iterations and used a smaller source length than HGCLR, which allowed HBGL to achieve similar inference speeds compared to HGCLR. Specifically, HGCLR achieves 1.02$\times$ to 1.10$\times$ inference speedups over HBGL on three datasets.

\section{Acknowledgments}
The research work is supported by the National Key Research and Development Program of China under Grant No. 2019YFA0707204, the National Natural Science Foundation of China under Grant Nos. 62276015, 62176014, the Fundamental Research Funds for the Central Universities.

\bibliography{custom}
\bibliographystyle{acl_natbib}

\appendix



\end{document}